\documentclass{article}

\usepackage{arxiv_default}
\usepackage[utf8]{inputenc}
\usepackage[T1]{fontenc}
\usepackage{hyperref}
\usepackage{url}
\usepackage{booktabs}
\usepackage{amsfonts}
\usepackage{amsmath}
\usepackage{amssymb}
\usepackage{nicefrac}
\usepackage{microtype}
\usepackage{xcolor}
\usepackage{algorithm}
\usepackage{algorithmic}
\usepackage{multirow}
\usepackage{graphicx}
\usepackage{subcaption}
\usepackage{placeins}
\usepackage{amsthm}
\providecommand{\mathbbm}[1]{\mathbf{#1}}
\usepackage{tikz}
\usepackage{pifont}
\usepackage{enumitem}

\newtheorem{remark}{Remark}

\newcommand{\mask}{\texttt{[M]}}
\newcommand{\MtoT}{\textrm{M2T}}
\newcommand{\TtoT}{\textrm{T2T}}
\newcommand{\TtoM}{\textrm{T2M}}
\newcommand{\Sstuck}{\mathcal{S}_{\mathrm{stuck}}}
\newcommand{\xold}{x_i^{\mathrm{old}}}
\newcommand{\xnew}{x_i^{*}}

\title{Remask, Don't Replace: Token-to-Mask Refinement in Diffusion Language Models}

\author{
  Lin Yao$^{1,2}$ \\
  $^1$School of Computer Science, Shanghai Jiao Tong University, Shanghai, 200240, China \\
  $^2$Zhongguancun Academy, Beijing, 100097, China \\
  \texttt{lin.yao@sjtu.edu.cn}
}

\begin{document}

\maketitle
%

\begin{abstract}
Diffusion language models (dLLMs) generate text through iterative denoising, filling multiple masked positions at each step.
Positions filled in the same step are predicted without conditioning on one another's newly filled values and can therefore be mutually inconsistent; once retained, these inconsistencies become context for later predictions.
We introduce \emph{Token-to-Mask} (T2M), a training-free inference-time correction method that identifies low-confidence positions using the model's probability of the current token, remasks them, and reconstructs them in later denoising steps.
On dLLMs equipped with correction mechanisms, a single T2M configuration transfers across tasks and models without retuning and broadly improves task metrics over each model's native correction mechanism.
In controlled experiments, we decompose correction methods into a detector that identifies suspicious tokens and an action that determines how to revise them.
Holding the detector fixed, remasking yields higher task metrics than replacement; across the tested detector--action combinations, current-token-probability detection paired with remasking performs best.
Compared with direct editing, T2M converts additional inference compute into performance gains more effectively and, on most tasks, retains a sequential-step advantage over autoregressive token-by-token decoding.
\end{abstract}

\section{Introduction}
\label{sec:intro}

Diffusion language models (dLLMs) generate text by starting from a fully masked sequence and iteratively revealing tokens.
Unlike autoregressive models, which generate tokens one by one in a fixed left-to-right order, dLLMs determine the reveal order using model confidence, such as token probabilities,
and typically commit multiple positions in parallel at each step~\citep{austin2021d3pm,sahoo2024mdlm,lou2024sedd}.
This parallelism can produce the same-length response in far fewer sequential decoding steps.
The LLaDA family~\citep{nie2025llada,nie2026llada21} further shows that dLLMs can be competitive with autoregressive models of comparable scale, so parallel decoding need not imply a large quality loss.
Even so, parallel commitment introduces a distinctive source of error propagation.
Tokens revealed in the same step are predicted from the same context, but without conditioning on each other's newly revealed values,
and may therefore be mutually inconsistent~\citep{kang2025parallelbench};
once committed, such inconsistencies are preserved and treated as context for subsequent token predictions unless an explicit revision mechanism intervenes~\citep{nie2026llada21}.

To mitigate this propagation, some models add a post-fill correction stage over already-visible tokens~\citep{nie2026llada21,wang2025remdm,zhai2026core,remedi2025,kim2025prism},
while others (e.g., LLaDA-8B~\citep{nie2025llada} and Dream~\citep{ye2025dream}) keep pure fill decoding with no such stage.
When a correction stage is included, existing systems mainly choose between two \emph{actions}.
\emph{Edit} (overwrite) replaces a visible token with another vocabulary token under the current context---
as in LLaDA2.1's Token-to-Token (\TtoT{}) editing~\citep{nie2026llada21} and CORE's perturbed-context resampling~\citep{zhai2026core}.
\emph{Remask} resets a visible token to \mask{} and lets later fill steps re-predict it.
ReMDM remasks visible tokens at random with a time-dependent probability, without targeting particular errors~\citep{wang2025remdm}.
RemeDi~\citep{remedi2025} ranks every position in the current block by the model's confidence, keeps the most confident $k$, remasks the rest, and increases $k$ over denoising steps.

\begin{figure}
    \centering
    \includegraphics[width=0.98\textwidth]{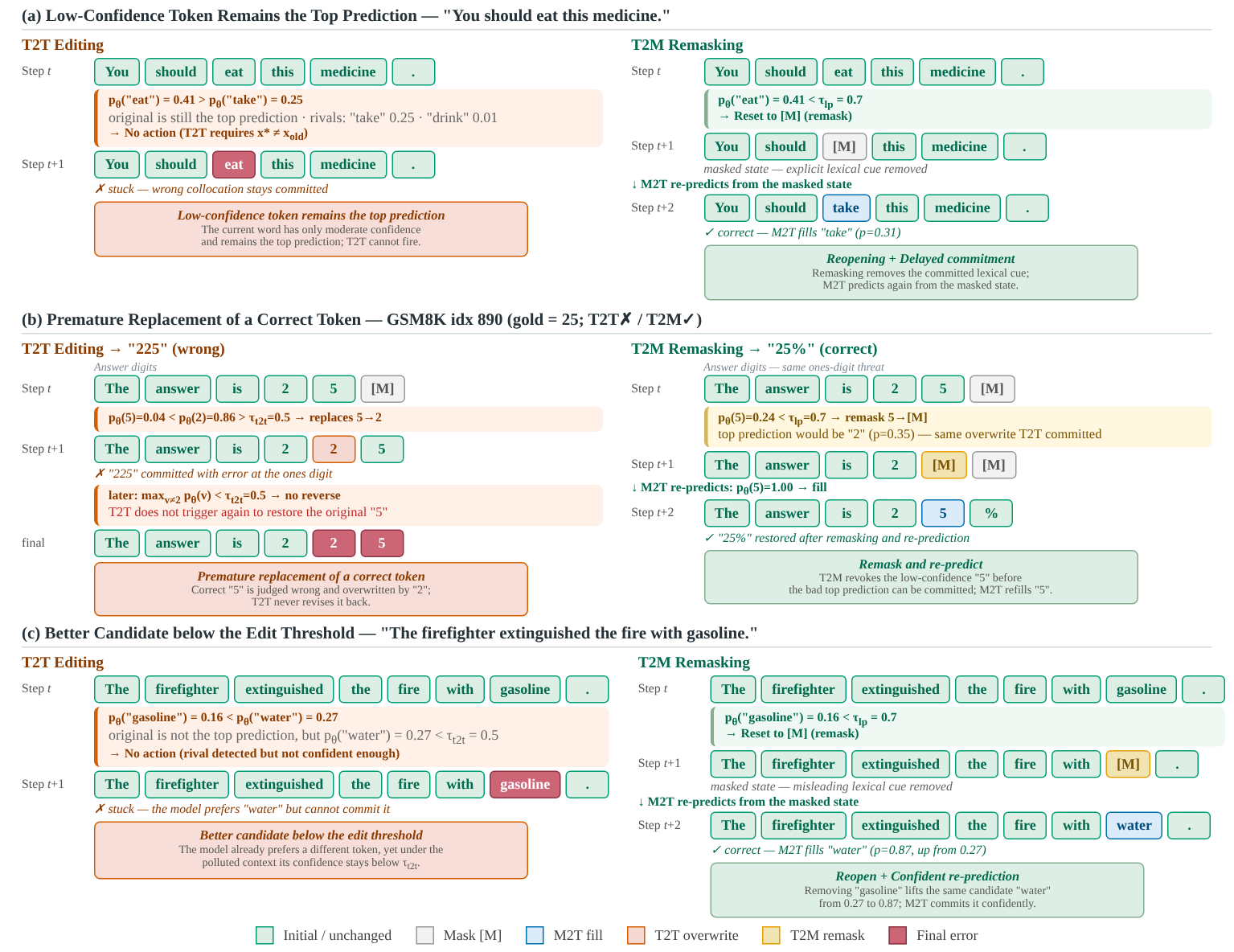}
\caption{
Three edit failure modes and \TtoM{} recovery on LLaDA2.1-mini:
\textbf{(a)}~low-confidence token remains the top prediction,
\textbf{(b)}~premature replacement of a correct token,
\textbf{(c)}~better candidate below the edit threshold.
Rows proceed from step $t$ to the final output.
Panels~(a) and~(c) use hand-constructed sentences and directly read the model's token probabilities.
Panel~(b) visualizes token changes observed while generating the answer to a GSM8K example.
}
    \label{fig:overview}
    \vspace{-0.5em}
  \end{figure}

\begin{figure}[t]
    \centering
    \includegraphics[width=\textwidth]{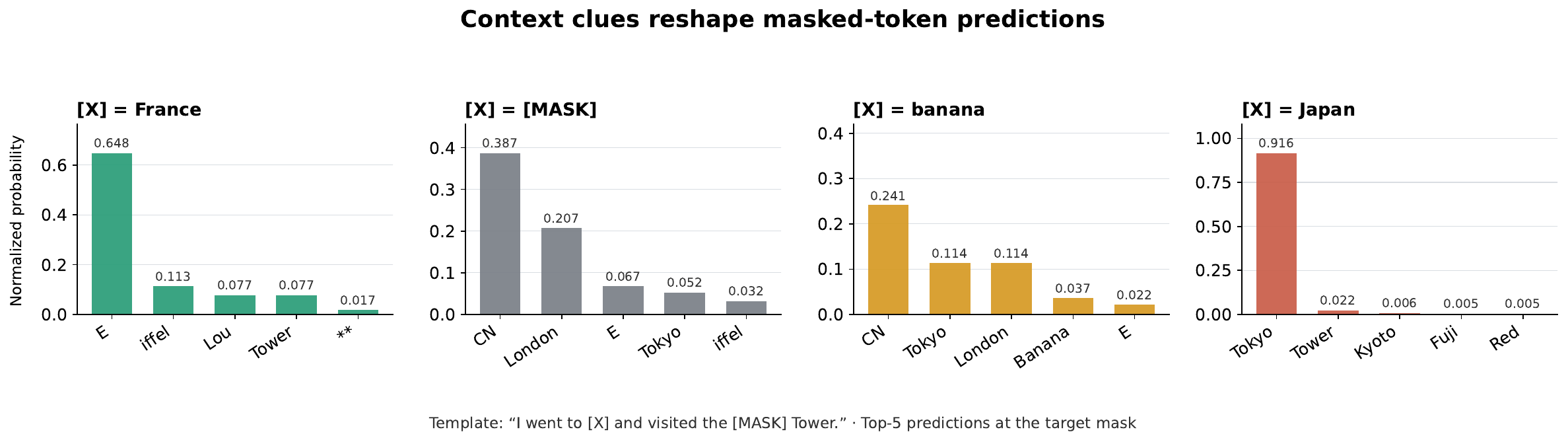}
    \caption{
    \textbf{Controlled context intervention on LLaDA2.1-mini.}
    For each of four context states at $[X]$ in ``I went to $[X]$ and visited the \mask{} Tower,''
    we plot the normalized top-5 distribution at the target mask.
    }
    \label{fig:signal}
\end{figure}

Edit typically fuses detecting an error with immediately overwriting it.
LLaDA2.1's \TtoT{} rule is a representative case:
it overwrites a committed token only when the model's top prediction under the current context differs from it \emph{and} that prediction's probability exceeds $\tau_{\text{t2t}}$.
This trigger produces the three failure modes in Fig.~\ref{fig:overview}.
In Fig.~\ref{fig:overview}(a), the current word has only moderate confidence and remains the top prediction, so the edit never fires.
In Fig.~\ref{fig:overview}(b), a correct token is overwritten because a confident but incorrect top prediction clears $\tau_{\text{t2t}}$.
In Fig.~\ref{fig:overview}(c), the model already prefers a different token, yet that candidate's probability stays below $\tau_{\text{t2t}}$, so the edit again never fires.
Across these cases, remasking permits later reconstruction without first proposing a confident replacement.
The traces therefore motivate our hypothesis that separating detection from immediate replacement is beneficial.
The fused trigger is not the only issue; overwrite as an action has two further drawbacks.
First, edit can create context pollution.
Fig.~\ref{fig:signal} illustrates this possibility: concrete context tokens can sharply reshape the prediction at another masked position, whereas masking the clue removes that token-specific signal and leaves the prediction to the remaining context and model prior.
Because edit must commit a concrete token rather than a masked placeholder, a wrong commitment remains an in-vocabulary, semantically loaded cue that can steer other positions toward continuations consistent with that cue,
whereas a \mask{} removes that explicit lexical cue and returns the position to the masked state seen during denoising.
Second, edit faces a training--inference noise mismatch:
edit streams are typically trained to recover from \emph{uniformly random} token perturbations injected into the input.
At inference time, however, the model's own mistakes are usually semantically plausible and locally coherent with their neighbours (Fig.~\ref{fig:overview}).
The editor is therefore asked to handle a kind of structured error it was not trained on, an instance of the broader training--inference divide identified for masked dLLMs~\citep{he2025mdpo}.

Remasking avoids forcing an immediate replacement, but existing remasking methods leave their own gaps.
ReMDM remasks at random: wrong commitments need not be remasked, while correct ones may be remasked and then refilled incorrectly.
RemeDi's keep-the-top-$k$ rule is likewise not a per-position error test: a high-confidence commitment can still fall outside the cutoff and be remasked, while a low-confidence one can remain inside top-$k$ and escape;
remasking is governed by cross-position rank rather than checking each position for error.

These gaps motivate a repair that keeps remask as the action but targets it using the same fill head that produces \MtoT{} predictions.
We propose \emph{Token-to-Mask} (\TtoM{}) refinement:
a token flagged by a simple confidence test on the model's own logits (e.g., its probability falling below a threshold $\tau_{\text{t2m}}$) is reset to \mask{} and re-predicted by the standard Mask-to-Token (\MtoT{}) fill rule like any other masked position.
The procedure is purely inference-time:
when the model already has a correction stage, it is a drop-in replacement for that stage.
The paper makes the following contributions:
\begin{itemize}[nosep,leftmargin=*]
\item We propose a training-free \TtoM{} rule guided by token probability: commitments whose probability falls below a threshold are reopened as \mask{}, decoupling detection from immediate replacement.
\item We demonstrate task generality with one fixed configuration across 17 benchmarks spanning math, code, science, reasoning, and instruction following.
\item We demonstrate transfer across models and correction mechanisms with one fixed configuration and no retuning: \TtoM{} replaces the native correction mechanism and improves performance on multiple models that already have a correction stage.
\item By separating detector from action, we find that remasking outperforms replacement under every matched detector on a controlled dataset, and moreover that probability-guided remasking is a strong combination.
\end{itemize}

\section{Related Work}
\label{sec:related}

\paragraph{Diffusion large language models.}
We briefly trace dLLMs from early diffusion models to current large-scale systems.
Diffusion generative modeling first appeared for continuous data, especially images~\citep{ho2020ddpm,song2021scorebased}.
D3PM~\citep{austin2021d3pm} then extended the framework to discrete tokens and singled out absorbing-state (mask) corruption as best suited to text.
A series of works then progressively simplified this formulation, including SEDD~\citep{lou2024sedd}, MDLM~\citep{sahoo2024mdlm}, and MD4~\citep{shi2024md4}.
RADD~\citep{ou2025radd} further showed that the predictor depends only on the unmasked context and not on the diffusion time $t$,
so a standard decoder-only Transformer~\citep{vaswani2017attention} suffices.
The framework has since been scaled up along two routes: LLaDA~\citep{nie2025llada} trains an 8B dLLM from scratch,
while Dream~\citep{ye2025dream} and DiffuLLaMA~\citep{gong2025diffullama} obtain comparable models by continued pretraining from autoregressive checkpoints;
industrial systems such as Mercury~\citep{inception2025mercury} push the same recipe to production scale.
Beyond scaling the fill model, a growing line of work adds a post-fill correction stage to repair errors that arise when multiple positions are committed in parallel~\citep{kang2025parallelbench,nie2026llada21,wang2025remdm,zhai2026core,remedi2025,kim2025prism}.
These strategies mainly take one of two actions---overwrite a committed token, or reopen it as \mask{}---which we review next.

\paragraph{Edit-based correction strategies.}
Edit overwrites an already-visible token with another vocabulary token under the current context.
LLaDA2.1~\citep{nie2026llada21} runs Token-to-Token (\TtoT{}) editing after each fill step:
every non-prompt visible position is re-scored, and the current token is replaced only when the model's top prediction differs from it and that prediction's probability clears a threshold $\tau_{\text{t2t}}$.
CORE~\citep{zhai2026core} instead selects which tokens to revise by measuring sensitivity to masked-context perturbations and resamples replacements from those perturbed contexts;
detection therefore relies on auxiliary probes rather than the logits on the actual generation trajectory, but the correction action remains overwrite.

\paragraph{Remask-based correction strategies.}
Remask resets an already-visible token to \mask{} and lets later fill steps re-predict it.
ReMDM~\citep{wang2025remdm} applies a uniform remasking probability $\sigma_t$ on each visible token.
RemeDi~\citep{remedi2025} re-ranks every position in the current block at each step by a confidence score---the fill-head logit of the committed token if the position is visible, otherwise of the argmax fill candidate---keeps only the top $K_n$ (with $K_n$ rising over denoising steps toward the block length), and remasks every position outside that cutoff.
PRISM~\citep{kim2025prism} fine-tunes a separate binary head that scores each visible token for remasking.
These methods avoid committing an immediate replacement, but either remask at random, fold remasking into a global unveiling order, or require additional training beyond the fill predictor.
Our approach stays in this family while using a training-free, position-wise detector from the shared fill-head logits on models that already include a correction stage (Section~\ref{sec:method}).

\section{Method}
\label{sec:method}

\subsection{Preliminaries}
\label{sec:background}

\paragraph{Mask-to-Token decoding.}
Masked dLLMs generate text through iterative Mask-to-Token (\MtoT{}) denoising~\citep{nie2025llada,ye2025dream,sahoo2024mdlm}.
The positions to be generated are initialised as \mask{} tokens.
At each denoising step, one forward pass produces a vocabulary distribution at every position that still holds \mask{};
the decoder then selects a subset of predictions according to its update rule and writes their token values into the corresponding positions.
The filled tokens become context for subsequent predictions.
The process repeats until no \mask{} remains.

\paragraph{Correction mechanisms.}
Parallel commitments are predicted from the same context without conditioning on one another and can therefore be mutually inconsistent~\citep{kang2025parallelbench}.
Some models therefore revise tokens after they have been filled.
We take LLaDA2.1's Token-to-Token (\TtoT{}) editing as an example~\citep{nie2026llada21}:
after each \MtoT{} step, every non-mask, non-prompt position is re-evaluated and replaced when its argmax prediction exceeds a threshold:
\begin{equation}
\begin{aligned}
    x_i^{*}
    &= \arg\max_{v \in \mathcal{V}} p_\theta(v \mid \mathbf{z}_i),
    \quad
    p_i^{*}
    = \max_{v \in \mathcal{V}} p_\theta(v \mid \mathbf{z}_i), \\
    x_i &\leftarrow
    \begin{cases}
        x_i^{*} & \text{if } p_i^{*} > \tau_{\mathrm{t2t}} \;\text{and}\; x_i^{*} \neq \xold, \\
        \xold   & \text{otherwise.}
    \end{cases}
\end{aligned}
\label{eq:t2t}
\end{equation}
Here $x_i$ is the token at position $i$ with current value $\xold$, $\mathbf{z}_i$ is the sequence state used to predict $i$, $\mathcal{V}$ is the vocabulary, $x_i^{*}$ and $p_i^{*}$ are the argmax token and its probability under $p_\theta(\cdot\mid\mathbf{z}_i)$, and $\tau_{\mathrm{t2t}}\in(0,1)$ is the editing threshold.
Appendix~\ref{app:background} reviews the discrete diffusion background in more detail.

\subsection{Token-to-Mask Remasking}
\label{sec:t2m}

\paragraph{Overview.}

\TtoM{} is a post-fill correction mechanism that replaces an existing correction rule, such as LLaDA2.1's \TtoT{}.
The model and the \MtoT{} fill rule stay unchanged.
At every non-mask, non-prompt position $i$, a detector judges whether the committed token remains trustworthy, and a flagged token is reset to \mask{}:
\begin{equation}
\begin{aligned}
    d_i
    &= \mathcal{D}_{\mathcal{S}}(i, \mathbf{z}_i, \xold, \tau_{\mathrm{t2m}}),
    \quad d_i \in \{0,1\}, \\
    x_i &\leftarrow
    \begin{cases}
        \mask{} & \text{if } d_i=1, \\
        \xold   & \text{otherwise.}
    \end{cases}
\end{aligned}
\label{eq:t2m}
\end{equation}
where $d_i\in\{0,1\}$ is the per-position remask indicator, $\mathcal{D}_{\mathcal{S}}$ is the detector under strategy $\mathcal{S}$, $\xold$ is the current token at position $i$,
and $\tau_{\mathrm{t2m}}$ is the strategy-specific remask threshold.
Algorithm~\ref{alg:t2m} specifies the full \TtoM{} procedure.
It runs after each \MtoT{} step.
On a model with a correction stage, \TtoM{} replaces that stage (\TtoT{} editing on LLaDA2.1);
on a model without one, the same rule is inserted after each fill step (Section~\ref{sec:crossmodel}).

\paragraph{Error detection.}

We consider three instantiations of the detector $\mathcal{D}_{\mathcal{S}}$, each built around a different signal:
\begin{align}
    d_i^{\mathrm{lp}}
    &= \mathbbm{1}\!\left\{p_\theta(\xold \mid \mathbf{z}_i) < \tau_{\mathrm{lp}}\right\},
    \label{eq:low_prob} \\[2pt]
    d_i^{\mathrm{tr}}
    &= \mathbbm{1}\!\left\{p_\theta(\xnew \mid \mathbf{z}_i) > \tau_{\mathrm{tr}}
       \text{ and } \xnew \neq \xold\right\},
    \quad \xnew \!=\! \arg\max_{v \in \mathcal{V}} p_\theta(v \mid \mathbf{z}_i),
    \label{eq:t2t_remask} \\[2pt]
    d_i^{\mathrm{ld}}
    &= \mathbbm{1}\!\left\{p_\theta^{(t-1)}(\xold \mid \mathbf{z}_i^{(t-1)})
       - p_\theta^{(t)}(\xold \mid \mathbf{z}_i^{(t)}) > \tau_{\mathrm{ld}}\right\}.
    \label{eq:logit_diff}
\end{align}
\begin{itemize}[nosep,leftmargin=*]
\item \textbf{\textsc{LowProb}} (Eq.~\ref{eq:low_prob}).
Remask whenever the model's probability for the currently committed token falls below threshold, with no dependence on any replacement candidate.
\item \textbf{\textsc{T2T-Remask}} (Eq.~\ref{eq:t2t_remask}).
Reuse \TtoT{}'s trigger and swap its action for remasking, isolating the effect of ``remask vs.\ replace'' under a fixed detector.
\item \textbf{\textsc{LogitDiff}} (Eq.~\ref{eq:logit_diff}).
Track how the model's confidence in $\xold$ evolves between consecutive iterations and fire on a probability drop.
A falling confidence indicates the converging neighbourhood has stopped supporting the token;
a rising one indicates it has been corroborated.
The signal depends on the denoising trajectory rather than a single snapshot;
\textsc{LogitDiff} abstains at the first iteration of a decoding window.
\end{itemize}

\paragraph{Safety caps.}

Remasking can in principle oscillate: a position is remasked, refilled with a similar token, remasked again.
Two caps prevent this.
A per-position budget $C_{\max}$ (default 3) limits how many times a single position may be remasked,
and a per-step ratio cap $\rho_{\max}$ (default $0.5$) limits the fraction of editable positions that can be remasked in one \MtoT{} iteration.
When the ratio cap binds, the least-confident positions are selected first.

\begin{algorithm}[!t]
\caption{Token-to-Mask (\TtoM{}) Remasking. Runs after each \MtoT{} step, in place of any existing correction stage.}
\label{alg:t2m}
\begin{algorithmic}[1]
\REQUIRE Current tokens $\mathbf{z}=(z_1,\ldots,z_B)$, positions $\mathcal{F}$ filled by the immediately preceding \MtoT{} step, model $p_\theta$, mask id \mask{}, detector strategy $\mathcal{S}$, threshold $\tau_{\mathrm{t2m}}$, per-position remask counts $\{c_i\}_{i=1}^{B}$ (initialized to $0$, persisted across \MtoT{} iterations), per-position budget $C_{\max}$, per-step ratio cap $\rho_{\max}$
\STATE $\mathcal{E} \leftarrow \{i : z_i \neq \mask{},\ i \notin \text{prompt},\ i \notin \mathcal{F}\}$ \hfill $\triangleright$ \textbf{editable positions}: previously filled, non-prompt
\STATE $\mathcal{R} \leftarrow \{i \in \mathcal{E} : \mathcal{D}_{\mathcal{S}}(i, \mathbf{z}_i, \xold, \tau_{\mathrm{t2m}})=1 \text{ and } c_i < C_{\max}\}$ \hfill $\triangleright$ \textbf{remask set}: flagged by the detector and under budget
\IF{$|\mathcal{R}| > \rho_{\max} \cdot |\mathcal{E}|$}
    \STATE $k \leftarrow \max\!\left(1,\lfloor \rho_{\max} \cdot |\mathcal{E}| \rfloor\right)$ \hfill $\triangleright$ ratio cap binds: keep top-$k$ most suspicious
    \STATE $\mathcal{R} \leftarrow$ top-$k$ positions in $\mathcal{R}$ ranked by $\mathcal{S}$
\ENDIF
\FOR{each $i \in \mathcal{R}$}
    \STATE $z_i \leftarrow \mask{}$; \quad $c_i \leftarrow c_i + 1$ \hfill $\triangleright$ reset flagged positions to \mask{}
\ENDFOR
\STATE \textbf{return} updated $\mathbf{z}$
\end{algorithmic}
\end{algorithm}

\section{Experiments}
\label{sec:experiments}
\subsection{Experimental Setup}
\label{sec:setup}

\paragraph{Models and decoding.}
LLaDA2.1-mini~\citep{nie2026llada21} (16B mixture-of-experts;
Appendix~\ref{app:models}) is the backbone for the 17-benchmark suite, the mechanism comparisons, and all ablations;
cross-model runs further include LLaDA2.1-flash (100B mixture-of-experts)~\citep{nie2026llada21}, RemeDi-Instruct (9B)~\citep{remedi2025}, Dream-7B (7B)~\citep{ye2025dream},
and LLaDA-8B-Base (8B)~\citep{nie2025llada}.
Both LLaDA2.1 models are served by SGLang dLLM~\citep{zheng2024sglang} under official Q~Mode defaults (fill $\tau_{\text{m2t}}{=}0.7$, edit $\tau_{\text{t2t}}{=}0.5$, block $B{=}32$, post-fill edit window $\le 16$, temperature $0$).
The correction practices under comparison are as follows.
\emph{Native baselines:} \TtoT{} editing on both LLaDA2.1 models;
RemeDi-Instruct's released top-$k$ remask under its released generator;
pure \MtoT{} (no correction stage) on Dream-7B and LLaDA-8B-Base under their released generators (Dream-7B:
temperature $0.1$, top-$p{=}0.9$, length/steps $256/256$;
LLaDA-8B-Base: length/steps/block $256/256/256$).
\emph{Our method:} \TtoM{} remasking (Section~\ref{sec:method}) replaces the native correction action when one exists,
and is inserted after each fill step when none does.
\emph{Additional baselines on LLaDA2.1-mini only:} ReMDM-style random remasking~\citep{wang2025remdm} ($\sigma\in\{0.02,0.05,0.10\}$, with higher-budget references reported separately) and CORE~\citep{zhai2026core} (gate $0.3$, interval $8$/$2$;
Appendix~\ref{app:core}), both ported onto the same Q~Mode loop.
Full per-model details are in Appendix~\ref{app:models}.

\paragraph{\TtoM{} configuration.}
Unless stated otherwise, \TtoM{} uses \textsc{LowProb} with $\tau_{\mathrm{lp}}{=}0.7$, $C_{\max}{=}3$, $\rho_{\max}{=}0.5$, selected on the GSM8K sweep of Section~\ref{sec:ablation} and then held fixed.
LLaDA2.1-flash reuses this setting without retuning;
RemeDi-Instruct uses the same $\tau_{\mathrm{lp}}{=}0.7$ with threshold-\MtoT{} fill $0.7$ in place of released top-$k$;
Dream-7B and LLaDA-8B-Base (no native correction stage) insert the same \textsc{LowProb} rule ($\tau_{\mathrm{lp}}{=}0.7$, $C_{\max}{=}3$, $\rho_{\max}{=}0.5$) after each fill step.

\paragraph{Benchmarks and protocols.}
The task-generality evaluation covers 17 benchmarks spanning math, code, science/knowledge, general reasoning,
and instruction following under paired task-standard protocols~\citep{ma2025dinfer,opencompass2023} (per-task split, prompt, and scorer in Appendix~\ref{app:benchmarks}).
The generation budget is $16{,}384$ tokens for the LLaDA2.1-mini 17-benchmark suite (TriviaQA: $50$ tokens).
On GSM8K, Table~\ref{tab:main} and Fig.~\ref{fig:design_ab} use $16{,}384$;
the 53-configuration sweep (each run on all 1,319 examples), the LLaDA2.1-mini block of Table~\ref{tab:mechanisms}, the top-$k$ ladder, and LLaDA2.1-flash use $4{,}096$ tokens.
At $4{,}096$ tokens, the cap binds for $<1\%$ of samples; all cells within a comparison share the same budget.
The GSM8K cross-model cells use each model's released prompt, budget, and scorer.
Every cell is the full split;
paired comparisons share weights, prompts, budget, engine, and scorer.

\paragraph{Metrics.}
Task scores follow each benchmark's official metric---accuracy, exact match, or pass@1---under the corresponding scorer (Appendix~\ref{app:benchmarks});
we also report mean generated length.
For compute we additionally report the \textbf{number of forward evaluations per generated token (NFE/token)}, measured per sample at batch size 1.
The value $1.0$ is one forward per generated token, the cost of autoregressive decoding without speculation.
Both edit and remask reuse that fill forward;
remask's extra cost is the later fill steps that close the reopened \mask{}s.
Wall-clock latency is in Appendix~\ref{app:latency};
reproduction notes in Appendix~\ref{app:repro}.

\subsection{Task Generality of \TtoM{}}
\label{sec:task}

\begin{table}
\centering
\caption{
Task generality of \TtoM{} on LLaDA2.1-mini (paired task-standard protocols).
$^\dagger$ denotes the GSM8K selection benchmark; the same configuration is used without per-task retuning on the other 16 tasks.
Metrics are accuracy, exact match (EM), or first-sample pass rate (pass@1), in percent.
Len.\ is mean generated length; Len.\ and NFE/tok.\ are \TtoT{}\,/\,\TtoM{} (AR $=1.0$).
}
\label{tab:main}
\vspace{4pt}
\small
\renewcommand{\arraystretch}{1.0}
\begin{tabular}{llrccccc}
\toprule
\textbf{Category} & \textbf{Benchmark} & $n$ & \textbf{\TtoT{}} & \textbf{\TtoM{}} & \textbf{$\Delta$} & \textbf{Len.} & \textbf{NFE/tok.} \\
\midrule
\multirow{5}{*}{Math}
  & GSM8K$^\dagger$  & 1319 & 89.61 & \textbf{91.21} & $+1.59$ & 296 / 292 & 0.26 / 0.58 \\
  & OlympiadBench    &  674 & 58.16 & \textbf{58.75} & $+0.59$ & 4440 / 4427 & 0.26 / 0.62 \\
  & GSM-Plus         & 10552 & 81.37 & \textbf{81.79} & $+0.42$ & 335 / 344 & 0.32 / 0.84 \\
  & Omni-MATH        & 4428 & 40.31 & \textbf{40.42} & $+0.11$ & 5452 / 5426 & 0.27 / 0.66 \\
  & CMATH            & 1098 & 92.26 & \textbf{92.35} & $+0.09$ & 301 / 302 & 0.32 / 0.77 \\
\midrule
\multirow{2}{*}{Code}
  & HumanEval+       &  164 & 76.83 & \textbf{80.49} & $+3.66$ & 353 / 335 & 0.14 / 0.29 \\
  & MBPP+            &  378 & 68.78 & \textbf{70.11} & $+1.33$ & 268 / 273 & 0.18 / 0.37 \\
\midrule
\multirow{4}{*}{Science/Knowl.}
  & GPQA-Diamond     &  198 & 43.94 & \textbf{45.96} & $+2.02$ & 4588 / 4629 & 0.41 / 1.12 \\
  & C-EVAL           & 1346 & 75.63 & \textbf{76.97} & $+1.34$ & 882 / 820 & 0.51 / 1.48 \\
  & MMLU-Pro         & 12032 & 63.41 & \textbf{63.56} & $+0.15$ & 1907 / 1921 & 0.37 / 0.97 \\
  & TriviaQA         & 17944 & 53.18 & \textbf{53.24} & $+0.06$ & 9 / 9 & 0.83 / 2.16 \\
\midrule
\multirow{5}{*}{Reasoning}
  & PIQA             & 1838 & 80.90 & \textbf{82.64} & $+1.74$ & 95 / 105 & 0.76 / 2.04 \\
  & OCNLI            & 2950 & 60.78 & \textbf{61.32} & $+0.54$ & 138 / 138 & 0.58 / 1.69 \\
  & DROP             & 9536 & 84.14 & \textbf{84.58} & $+0.44$ & 232 / 223 & 0.33 / 0.73 \\
  & BBH              & 6511 & \textbf{81.92} & 81.82 & $-0.10$ & 665 / 620 & 0.42 / 0.96 \\
  & HellaSwag        & 10042 & \textbf{79.50} & 79.38 & $-0.12$ & 251 / 250 & 0.55 / 1.50 \\
\midrule
Instruction
  & IFEval (prompt-strict) & 541 & 83.36 & \textbf{84.47} & $+1.11$ & 461 / 482 & 0.70 / 2.17 \\
\bottomrule
\end{tabular}
\end{table}

Table~\ref{tab:main} tests task generality under one shared \TtoM{} configuration; as noted in the caption, GSM8K is the configuration-selection benchmark and the same setting is used without per-task retuning elsewhere.
\TtoM{} improves the task metric on $15$ of $17$ benchmarks;
the two non-improvements stay within $0.12$ points of the baseline, while the largest increases are HumanEval+ ($+3.66$), GPQA ($+2.02$), and GSM8K ($+1.59$).
Mean generated lengths are reported to make any decoder-dependent length changes explicit.
\paragraph{Compute cost.}
\label{sec:cost}
These improvements incur additional compute.
Remasking reopens tokens as \mask{}, and later fill steps must close them:
relative to \TtoT{}, \TtoM{} uses $2.0$--$3.1\times$ as many forwards.
On GSM8K, both the forward count and measured latency increase by $2.2\times$; full per-task latency results are in Appendix~\ref{app:latency}.
Ten of seventeen tasks remain below the AR reference of $1.0$.
The other seven are either short generations (TriviaQA and PIQA, roughly $10$--$10^{2}$ tokens, where fixed remask overhead meets a small denominator) or tasks whose $2$--$3\times$ multiplier raises NFE/token above $1.0$ (GPQA, C-EVAL, IFEval, HellaSwag, OCNLI).
If NFE/token exceeds $1.0$, tightening the remask budget can recover a point below it.
We illustrate this on C-EVAL ($C_{\max}{=}1$, $\rho_{\max}{=}0.25$: $77.04$ at $0.86$ NFE/token) and do not apply the same budget tightening to the remaining tasks (Appendix~\ref{app:latency}).

\subsection{Cross-Model Transfer}
\label{sec:crossmodel}

\begin{table}[!htbp]
    \centering
\caption{
Cross-model study.
Each baseline uses that model's released generator;
\TtoM{} replaces native post-fill correction when present and is inserted after fill otherwise, with the remaining released settings fixed (Section~\ref{sec:setup}).
}
    \label{tab:crossmodel}
    \vspace{4pt}
    \small
    \begin{tabular}{lllccc}
    \toprule
    \textbf{Model} & \textbf{Correction training} & \textbf{Benchmark} & \textbf{Baseline} & \textbf{\TtoM{}} & \textbf{$\Delta$} \\
    \midrule
    LLaDA2.1-flash 100B~\citep{nie2026llada21} & \TtoT{} editing stream & GSM8K & 92.87 & \textbf{94.31} & $+1.44$ \\
    LLaDA2.1-flash 100B & \TtoT{} editing stream & MBPP+ & 76.46 & \textbf{77.78} & $+1.32$ \\
    LLaDA2.1-flash 100B & \TtoT{} editing stream & IFEval & 84.29 & \textbf{85.40} & $+1.11$ \\
    LLaDA2.1-flash 100B & \TtoT{} editing stream & C-EVAL & 85.36 & \textbf{86.03} & $+0.67$ \\
    LLaDA2.1-flash 100B & \TtoT{} editing stream & PIQA & 90.48 & \textbf{91.02} & $+0.54$ \\
    LLaDA2.1-flash 100B & \TtoT{} editing stream & CMATH & 94.08 & \textbf{94.35} & $+0.27$ \\
    RemeDi-Instruct~\citep{remedi2025} & top-$k$ remask & GSM8K & 86.28 & \textbf{87.04} & $+0.76$ \\
    Dream-7B~\citep{ye2025dream} & none & GSM8K & \textbf{79.53} & 79.23 & $-0.30$ \\
    LLaDA-8B-Base~\citep{nie2025llada} & none & GSM8K & 73.31 & \textbf{73.39} & $+0.08$ \\
    \bottomrule
    \end{tabular}
    \end{table}

\TtoM{} uses the configuration selected on LLaDA2.1-mini without retuning and is compared against each model's released generator (Table~\ref{tab:crossmodel}).
\emph{Targeted remasking shows positive transfer on models that already have a correction stage.}
On the six reported LLaDA2.1-flash tasks, \TtoM{} improves performance, with the largest gain on GSM8K ($+1.44$).
Dream-7B and LLaDA-8B-Base have no correction training; the gain is absent or not apparent.

\subsection{Detector and Action Comparison}
\label{sec:mechanisms}

Correction methods bind a detector to an action.
To compare the two independently, we cross detectors with both actions on full GSM8K and compare against native correction (Table~\ref{tab:mechanisms}).
Under the same detector, remasking outperforms replacement.
Overall, probability-guided remasking is the strongest combination: it outscores \TtoT{}, CORE, and random remasking, and it also surpasses native correction, including RemeDi's released method.

\begin{table}[!htbp]
\centering
\caption{
Detector $\times$ action on full GSM8K with a $4{,}096$-token generation budget.
Each cell reports accuracy @ NFE/token (AR $=1.0$).
Upper block: LLaDA2.1-mini;
bottom: RemeDi-Instruct under its released loop (Appendix~\ref{app:models}).
Unless a row notes otherwise, the remask budget is $C_{\max}{=}3$, $\rho_{\max}{=}0.5$.
}
\label{tab:mechanisms}
\vspace{4pt}
\small
\begin{tabular}{lcc}
\toprule
\textbf{Detector $\backslash$ action} & \textbf{Replace} & \textbf{Remask} \\
\midrule
None (pure \MtoT{} decoding, base) & \multicolumn{2}{c}{86.58 @ 0.283} \\
\midrule
Random, $\sigma{=}0.02$, $C{=}3$ (ReMDM-style) & --- & 87.11 @ 0.43 \\
Random, $\sigma{=}0.05$, $C{=}3$ & --- & 89.23 @ 0.98 \\
Random, $\sigma{=}0.10$, $C{=}3$ & --- & 89.76 @ 1.35 \\
Random, $\sigma{=}0.10$, $C{=}8$ & --- & 91.28 @ 2.88 \\
\midrule
\TtoT{} trigger & 89.54 @ 0.262 (official) & 89.76 @ 0.267 (\textsc{T2T-Remask}) \\
CORE sensitivity, interval 8 & 87.11 @ 0.496 (CORE) & 88.86 @ 0.457 \\
CORE sensitivity, interval 2 & 85.22 @ 0.642 & 89.39 @ 0.523 \\
\textsc{LowProb} & 89.39 @ 0.258 & $\mathbf{91.13}$ \textbf{@ 0.584} (\TtoM{}, default) \\
\textsc{LowProb}, $\tau{=}0.9$, $C{=}3$ & --- & $\mathbf{91.66}$ \textbf{@ 0.672} \\
\midrule
\multicolumn{3}{l}{\emph{On RemeDi-Instruct's released loop (Appendix~\ref{app:models}):}} \\
RemeDi released confidence score & 67.02 @ 1.08 & 86.28 @ 1.08 (released top-$K$) \\
\textsc{LowProb} & 39.73 @ 0.42 & $\mathbf{87.04}$ \textbf{@ 0.52} (\TtoM{}) \\
\bottomrule
\end{tabular}
\end{table}

\subsection{Ablation Studies}
\label{sec:ablation}

\begin{figure}[htbp]
    \centering
    \renewcommand{\thesubfigure}{\Alph{subfigure}}
    \begin{subfigure}{0.98\textwidth}
        \centering
        \includegraphics[width=\textwidth]{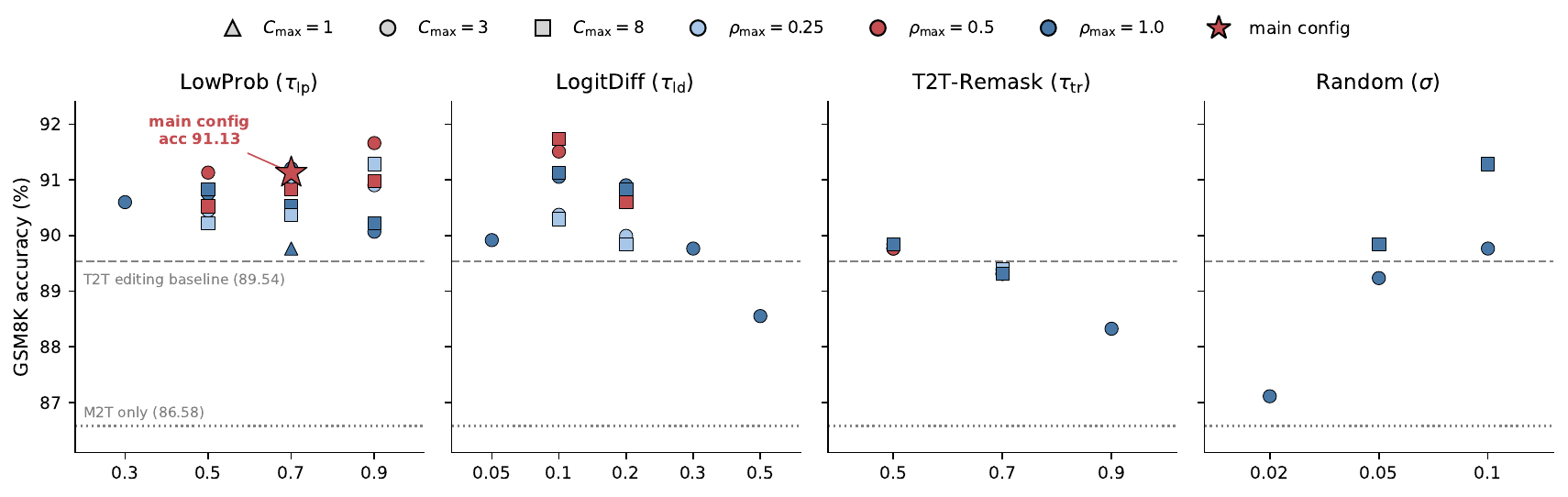}
        \caption{\textbf{Detectors under remasking.}
        Marker shape encodes $C_{\max}$, color encodes $\rho_{\max}$, and the star marks the default configuration.
        The dashed and dotted lines are the \TtoT{} and pure-\MtoT{} baselines at the shared $4{,}096$-token budget.}
        \label{fig:sweep}
    \end{subfigure}
    \vspace{0.6em}
    
    \begin{subfigure}{0.98\textwidth}
        \centering
        \includegraphics[width=\textwidth]{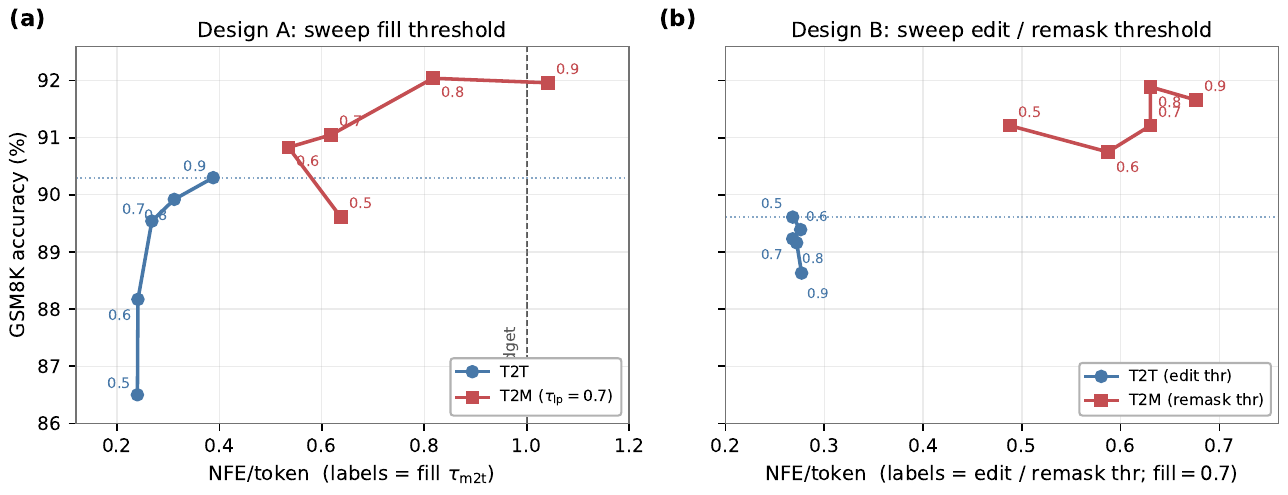}
        \caption{\textbf{Accuracy--compute frontiers.}
        Design~A sweeps the \MtoT{} fill threshold; Design~B fixes fill at $0.7$ and sweeps the correction threshold.
        Point labels are thresholds, the vertical dashed line is the AR reference of $1.0$, and the horizontal dotted line is the best tested \TtoT{} accuracy.}
        \label{fig:design_ab}
    \end{subfigure}
    \caption{\textbf{Detector ablation and compute scaling on GSM8K} (full 1319-problem set).
    Panel~(A) holds remasking fixed and compares detectors across thresholds and budgets;
    panel~(B) tests whether extra compute converts into accuracy.}
    \label{fig:ablation_compute}
    \end{figure}

Fig.~\ref{fig:sweep} holds remasking fixed and compares detectors---\textsc{LowProb}, \textsc{LogitDiff}, and the \TtoT{} trigger---plus an untargeted random control, sweeping each detector's threshold and remask budget on full GSM8K.
\textsc{LowProb} and \textsc{LogitDiff} both outscore \TtoT{}-triggered remasking and random remasking.
Every \textsc{LowProb} setting lies above \TtoT{} editing, so the gain is not an artifact of one operating point.
Swapping \TtoT{}'s overwrite for remasking leaves accuracy essentially unchanged, so the detector matters as well.
\textsc{LogitDiff} is a harder signal than \textsc{LowProb}---it reads the denoising trajectory rather than a single-token probability---and it does not buy additional accuracy.

\paragraph{Accuracy--compute frontier.}
\label{sec:pareto}
Fig.~\ref{fig:design_ab} tests whether extra compute converts into accuracy.
Design~A sweeps the fill threshold, and Design~B fixes fill while sweeping the edit or remask threshold.
Additional fill or more aggressive editing does not bring \TtoT{} to the \TtoM{} frontier; \TtoM{} turns the extra compute into higher accuracy.
The tested fill- and edit-threshold sweeps likewise do not close the gap on HumanEval+, MBPP+, C-EVAL, and PIQA (Appendix~\ref{app:design_ab}).
Taking accuracy and compute together from both panels, we use \textsc{LowProb} with $\tau_{\mathrm{lp}}{=}0.7$, $C_{\max}{=}3$, and $\rho_{\max}{=}0.5$ as the overall setting for the cross-task and cross-model evaluation.

Trajectory-level remask statistics and outcome-flip analyses are reported in Appendix~\ref{app:mechanism}.

\section{Conclusion}
\label{sec:conclusion}

We introduced Token-to-Mask (\TtoM{}), a training-free post-fill correction method for diffusion language models.
Rather than immediately replacing suspicious tokens, \TtoM{} identifies low-confidence positions from the probabilities of their current tokens, remasks them, and lets subsequent standard \MtoT{} steps reconstruct them.
Without task- or model-specific retuning, a single \TtoM{} configuration yields consistent gains across diverse tasks and multiple models equipped with post-fill correction, demonstrating cross-task and cross-model generality.
Decomposing correction into detector and action shows that, with the detector held fixed, remasking outperforms replacement; detector ablations further show that identifying which tokens to revise is equally important, and that current-token probability provides a simple and effective per-position signal.
Overall, by selectively reopening low-confidence positions and deferring reconstruction, \TtoM{} mitigates several failure modes of direct editing and converts additional inference compute into performance more effectively than editing; on most tasks, these gains retain the sequential-step advantage of dLLMs over autoregressive token-by-token decoding.

\paragraph{Limitations.}
\TtoM{} relies on the reconstruction capability induced by correction training; on fill-only models without such training, remasking yields little benefit.
\TtoM{} also requires additional denoising steps to refill remasked positions, resulting in $2.0$--$3.1\times$ the NFE of \TtoT{} and therefore higher inference cost.

\bibliographystyle{plainnat}
\bibliography{references}

\clearpage
\appendix
\section*{\centering\Large Appendix}
\setcounter{figure}{0}
\setcounter{table}{0}
\setcounter{equation}{0}
\setcounter{algorithm}{0}
\setcounter{definition}{0}
\setcounter{proposition}{0}
\setcounter{remark}{0}
\renewcommand{\thefigure}{S\arabic{figure}}
\renewcommand{\thetable}{S\arabic{table}}
\renewcommand{\theequation}{S\arabic{equation}}
\renewcommand{\theHequation}{S.\arabic{equation}}
\renewcommand{\thealgorithm}{S\arabic{algorithm}}
\renewcommand{\thedefinition}{S\arabic{definition}}
\renewcommand{\theproposition}{S\arabic{proposition}}
\renewcommand{\theremark}{S\arabic{remark}}

\section{Background}
\label{app:background}

\subsection{Diffusion Large Language Models}

A diffusion large language model learns to recover clean text from partially masked inputs~\citep{austin2021d3pm,ou2025radd,sahoo2024mdlm}. Given a clean sequence $\mathbf{x} = (x_1, \ldots, x_L)$, training samples a mask ratio $r$ and a corresponding masked index set $\mathcal{M} \subseteq \{1,\ldots,L\}$. The corrupted sequence $\mathbf{z}$ is defined by replacing $x_i$ with \mask{} for $i \in \mathcal{M}$ and leaving the remaining tokens unchanged.

The model $p_\theta(\mathbf{x} \mid \mathbf{z})$ is trained to predict all masked positions simultaneously via a cross-entropy loss:
\begin{equation}
    \mathcal{L}(\theta) = -\mathbb{E}_{\mathbf{x}, r, \mathcal{M}}\left[\frac{1}{|\mathcal{M}|} \sum_{i \in \mathcal{M}} \log p_\theta(x_i \mid \mathbf{z})\right].
    \label{eq:mdlm_loss}
\end{equation}

At inference time, generation iteratively fills a masked sequence while keeping the prompt visible. Each step commits a subset of predictions, which then become context for subsequent steps.

\section{Mechanistic and Idealized Analysis}
\label{app:theory}

\subsection{Detection--Replacement Coupling: The Stuck Set}
\label{app:stuck}

\TtoT{}'s trigger fuses two questions into a single confidence test: whether the committed token $\xold$ is wrong, and whether a sufficiently confident replacement exists. We formalise the resulting failure regime as the \emph{stuck set}, for a small confidence floor $\epsilon$:
\begin{equation}
    \Sstuck
    = \left\{i :
        p_\theta(\xold \mid \mathbf{z}_i) < \epsilon
        \;\text{and}\;
        \max_{v \in \mathcal{V}} p_\theta(v \mid \mathbf{z}_i) < \tau_{\text{t2t}}
      \right\}.
    \label{eq:stuck_main}
\end{equation}
The first clause picks out positions that any detector based on the committed token's own probability would catch whenever its threshold exceeds $\epsilon$; the second is precisely the negation of \TtoT{}'s trigger. Hence \TtoT{} cannot fire at any $i \in \Sstuck$, while the \TtoM{} \textsc{LowProb} strategy does whenever $\tau_{\mathrm{t2m}}>\epsilon$. Fig.~\ref{fig:overview}(a) shows a controlled instance of this regime: the committed ``eat'' (wrong collocation with ``medicine'') has only moderate probability ($0.41$) and remains the top prediction (best rival ``take'' at $0.25$), so \TtoT{}'s requirement $x_i^{*} \neq \xold$ fails. Fig.~\ref{fig:overview}(c) is the complementary case: the model already prefers a different token (``water'' at $0.27$ vs.\ ``gasoline'' at $0.16$), yet that rival also stays below $\tau_{\text{t2t}}$, so the confidence clause of the trigger fails. In both cases the token's own moderate probability still falls below a \textsc{LowProb} threshold. Remasking does not guarantee a correct re-prediction, but reopening the position is a necessary condition for one.

\subsection{Removing Token-Specific Cues: A More Detailed Account}
\label{app:purification}

A token's effect on every other position depends on what is written at that position: the model treats every committed token as evidence. This gives a diagnostic hierarchy of context signals.

\begin{remark}[Context signal types]
\label{rem:signal}
For predicting $x_i$, each context position $j \neq i$ provides one of three signal types:
\begin{itemize}[nosep,leftmargin=*]
    \item \textbf{Aligned} (the correct token): supports the intended continuation.
    \item \textbf{Masked} (\mask{}): removes an explicit lexical commitment and is a familiar state for the \MtoT{} predictor, but its effect remains context- and model-dependent.
    \item \textbf{Misleading} (a wrong but committed token): supplies a coherent but erroneous cue.
\end{itemize}
\end{remark}

These signal types explain the action difference without assuming that every mask is neutral. Replacement can be more beneficial when its argmax is correct, but riskier when that argmax is computed under erroneous context. Remasking instead removes a concrete vocabulary-level cue and returns the position to the model's familiar masked state.

At every denoising step, the prediction at position $i$ is conditioned on the values at all other positions, which can be decomposed as
\begin{equation}
    p_\theta(x_i \mid \mathbf{z}_i) = p_\theta\!\left(x_i \;\middle|\; \underbrace{\mathbf{z}_{\text{correct}}}_{\text{correct tokens}},\; \underbrace{\mathbf{z}_{\text{error}}}_{\text{erroneous tokens}},\; \underbrace{\mask{}, \ldots, \mask{}}_{\text{still unknown}}\right).
\end{equation}
The erroneous positions in $\mathbf{z}_{\text{error}}$ arise from two sources: an \MtoT{} step may commit an incorrect fill, or a \TtoT{} step may replace a token with a different but still incorrect one. In either case, the result is indistinguishable from a correct token to the model and is processed as informative context.

\TtoT{} writes the current argmax back into the sequence, even though that argmax may be computed under other erroneous commitments. \TtoM{} instead converts suspected misleading signals to \mask{} and defers their reconstruction, preventing a new vocabulary token from immediately reinforcing the same erroneous context. Remasking does not guarantee a correct refill; its benefit depends on the context becoming more informative before the position is committed again.

\section{Extended Comparison with Prior Remasking Methods}
\label{app:related}

Section~\ref{sec:related} summarises the qualitative contrast with prior remasking methods. This appendix gives an idealized comparison between targeted and random remasking under explicit assumptions.

\subsection{Targeted vs.\ Random Remasking: An Idealized Comparison}
\label{app:targeted_vs_random}

ReMDM~\citep{wang2025remdm} remasks each committed token independently with probability $\sigma_t$ at every reverse-diffusion step. To isolate the selection effect, consider one reverse step and fix $\sigma=\sigma_t$. Let there be $N$ committed (non-mask) positions, of which $N_c$ are correct (contribution $s_+ > 0$) and $N_e$ are erroneous (contribution $s_- < 0$). Under the idealized assumptions that per-position contributions are additive and that a remasked position contributes zero at this step, define the context quality as $Q = N_c s_+ + N_e s_-$. Under random remasking with per-token probability $\sigma$, each position is removed independently of correctness, so in expectation
\begin{equation}
    Q_{\text{random}}(\sigma) = (1 - \sigma)(N_c s_+ + N_e s_-).
    \label{eq:q_random}
\end{equation}
Increasing $\sigma$ eliminates erroneous signals, but in equal proportion destroys correct signals; there is no $\sigma^{*}$ at which the trade-off is uniformly favourable. Under targeted remasking with perfect detection, only the $N_e$ erroneous positions are removed, yielding
\begin{equation}
    Q_{\text{targeted}} = N_c s_+.
    \label{eq:q_targeted}
\end{equation}
Within this idealized model, the difference is non-negative for all $\sigma \in [0, 1]$:
\begin{equation}
    Q_{\text{targeted}} - Q_{\text{random}}(\sigma) = \underbrace{\sigma N_c s_+}_{\text{correct signals preserved}} + \underbrace{(1 - \sigma)(-N_e s_-)}_{\text{erroneous signals removed}} \ge 0.
    \label{eq:q_advantage}
\end{equation}
Both terms are non-negative, so under perfect detection and the assumptions above, targeted remasking preserves at least as much correct context and removes at least as much error signal as the random alternative. Actual masked states need not have zero effect, and with imperfect detection the conclusion additionally requires assumptions on detector precision; we do not claim matched-count dominance for the implemented detector.

\section{Model Details}
\label{app:models}

LLaDA2.1-mini~\citep{nie2026llada21} is a 20-layer mixture-of-experts diffusion language model with 16B non-embedding parameters: 256 routed experts with 8 active per token plus one shared expert, hidden size 2048, a 32k-token context window, and a 157k vocabulary. Mini and LLaDA2.1-flash (the 100B MoE sibling) are decoded with the SGLang dLLM backend~\citep{zheng2024sglang}, flash with 4-way tensor parallelism. Cross-model GSM8K baselines in Section~\ref{sec:crossmodel} each use that model's own released code, the GSM8K test set (1319), and its published decoding settings; \TtoM{} replaces native post-fill correction when present and is inserted after fill otherwise, with the remaining released settings fixed. Dream-7B~\citep{ye2025dream} uses $\mathtt{diffusion\_generate}$ under the instruct GSM8K protocol (max\_new\_tokens/steps $256/256$, temperature $0.1$, top-$p$ $0.9$, entropy); LLaDA-8B-Base~\citep{nie2025llada} uses the official $\mathtt{generate}$ under the OpenCompass base recipe (4-shot CoT; gen/steps/block $256/256/256$). RemeDi-Instruct~\citep{remedi2025} keeps the released prompt-KV / block denoising loop (Appendix~B.3.2 boxed MATH; gen $1024$, block $32$); the paired \TtoM{} leg swaps the released cumulative top-$k$ remask schedule for \textsc{LowProb} remasking under threshold-\MtoT{} fill.

\paragraph{RemeDi detector $\times$ recipe grid.} The bottom block of Table~\ref{tab:mechanisms} reports the substitution grid behind the RemeDi discussion of Section~\ref{sec:mechanisms}: detector (RemeDi's released fill-head confidence score vs.\ our \textsc{LowProb}) crossed with correction recipe (the released top-$K$ unveiling, \TtoM{} bounded remask, or argmax overwrite), on the released loop and full GSM8K. The \textsc{LowProb} rows also replace RemeDi's global top-$K$ fill schedule with threshold fill, so this block is not a one-variable action ablation. RemeDi's released decoder re-ranks \emph{every} position at each step and reveals only the top $K_n$ (rising linearly to the block length), so reopening falls out of the ranking---unbudgeted and aimed at no particular token. The per-sample counts behind the NFE/token figures: the released schedule modifies ${\sim}384$ positions per sample at $274$ forwards, \TtoM{} modifies ${\sim}112$ at $134$, and the two overwrite legs modify ${\sim}348$ at $270$ (released score) and ${\sim}90$ at $96$ (\textsc{LowProb}).

\section{Hyperparameter Sweep Details}
\label{app:ablation}

The default configuration (\textsc{LowProb}, $\tau_{\mathrm{lp}}{=}0.7$, $C_{\max}{=}3$, $\rho_{\max}{=}0.5$) is selected as a conservative accuracy--compute trade-off in the GSM8K selection and sensitivity sweep of Section~\ref{sec:ablation} and then held fixed for the reported transfer evaluation.

The sweep evaluates 53 (strategy, $\tau_{\mathrm{t2m}}$, $C_{\max}$, $\rho_{\max}$) configurations, each on the full GSM8K test set (1319 problems). All shared inference parameters follow the LLaDA2.1 Q~Mode defaults of Section~\ref{sec:setup}: \MtoT{} fill threshold $\tau_{\text{m2t}}{=}0.7$ (shared by every configuration, including the baseline), block length $B{=}32$, greedy decoding (temperature $0$), single-sample inference (batch size 1) on a single A100-80GB GPU per configuration, and the $4{,}096$-token generation budget discussed in Section~\ref{sec:ablation}. The denoising inner loop runs until convergence (all masks filled and no edits/remasks triggered), not for a fixed number of steps. The baseline uses unmodified \TtoT{} editing at $\tau_{\text{t2t}}{=}0.5$~\citep{nie2026llada21}.

The sweep covers three remask-specific hyperparameters:
\begin{enumerate}[nosep,leftmargin=*]
    \item The \emph{remask threshold} $\tau_{\mathrm{t2m}}$ governs how aggressively each strategy triggers remasking. The sign of the effect depends on the strategy (Table~\ref{tab:tau_dir}).
    \begin{table}[!t]
    \centering\small
    \vspace{-2pt}
    \caption{Effect of increasing $\tau_{\mathrm{t2m}}$ on remask volume per strategy.}
    \label{tab:tau_dir}
    \vspace{3pt}
    \begin{tabular}{lcl}
    \toprule
    \textbf{Strategy} & \textbf{Higher $\tau_{\mathrm{t2m}}$ $\to$} & \textbf{Sweep range} \\
    \midrule
    \textsc{LowProb}    & more remasking & $\{0.3, 0.5, 0.7, 0.9\}$ \\
    \textsc{T2T-Remask} & less remasking & $\{0.5, 0.7, 0.9\}$ \\
    \textsc{LogitDiff}  & less remasking & $\{0.05, 0.1, 0.2, 0.3, 0.5\}$ \\
    \bottomrule
    \end{tabular}
    \vspace{-6pt}
    \end{table}
    \item The \emph{per-position budget} $C_{\max} \in \{1, 3, 8\}$ limits how often any one position can be remasked.
    \item The \emph{ratio cap} $\rho_{\max} \in \{0.25, 0.50, 1.0\}$ caps the fraction of editable positions remasked in a single step; $\rho_{\max}=1.0$ corresponds to no cap.
\end{enumerate}
In addition, the untargeted random-remasking baseline (ReMDM-style, Section~\ref{sec:mechanisms}) is swept over the per-step remask probability $\sigma \in \{0.02, 0.05, 0.10\}$ under the same caps. The grid is non-exhaustive and contains 53 configurations in total; Fig.~\ref{fig:sweep} plots every evaluated configuration, so the sensitivity evidence is not restricted to the selected point. Separately from the remask sweep, Table~\ref{tab:mechanisms} includes one additional matched-budget run that keeps the \textsc{LowProb} detector ($\tau_{\mathrm{lp}}{=}0.7$, $C_{\max}{=}3$, $\rho_{\max}{=}0.5$) but restores a replace action.

\subsection{Multi-task Accuracy--Compute Frontiers}
\label{app:design_ab}

\begin{figure}[t]
    \centering
    \includegraphics[width=\textwidth]{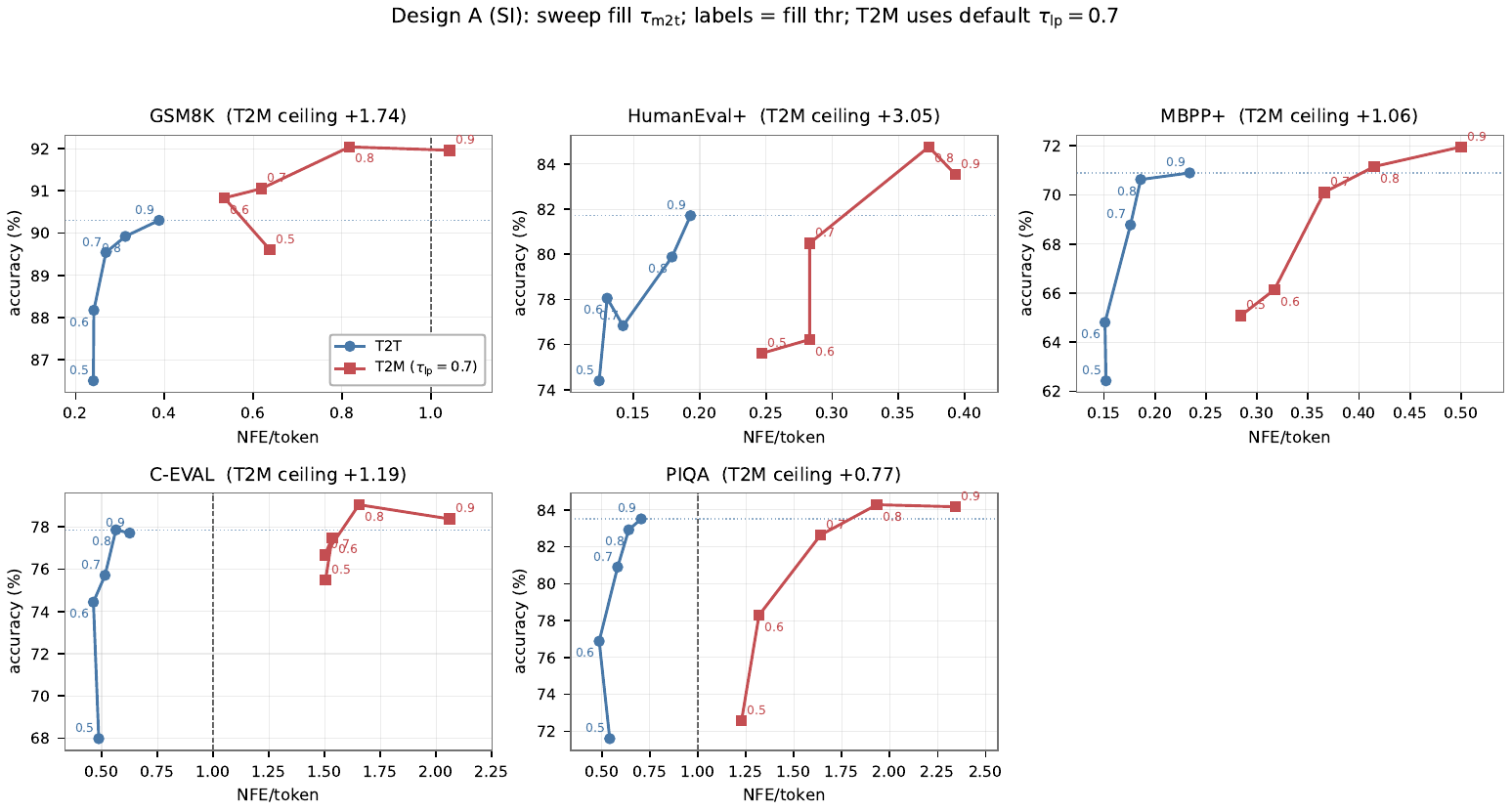}
    \caption{\textbf{Design~A on five benchmarks} (SI; $16{,}384$-token generation budget). Sweep \MtoT{} fill threshold $\tau_{\text{m2t}}\in\{0.5,\ldots,0.9\}$; $x$-axis is NFE/token and point labels are the fill threshold. \TtoM{} uses the default \textsc{LowProb} threshold $\tau_{\mathrm{lp}}{=}0.7$ ($C_{\max}{=}3$, $\rho_{\max}{=}0.5$). Horizontal dotted line: best tested \TtoT{} accuracy on that panel; vertical dashed line (when visible): AR reference $1.0$.}
    \label{fig:design_a_multitask}
\end{figure}

\begin{figure}[t]
    \centering
    \includegraphics[width=\textwidth]{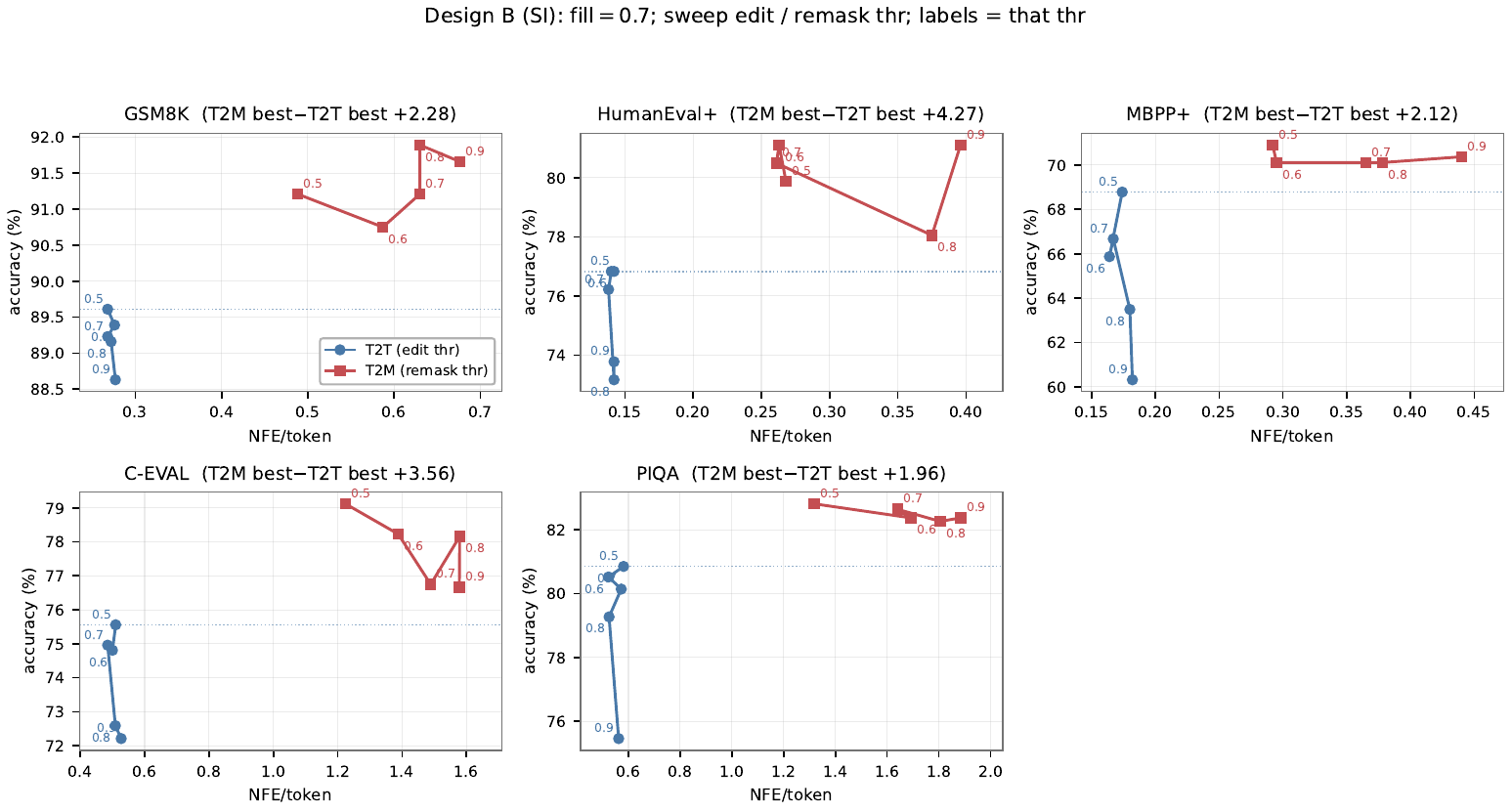}
    \caption{\textbf{Design~B on five benchmarks} (SI; $16{,}384$-token generation budget). Fill fixed at $0.7$; sweep the \TtoT{} edit threshold or the \TtoM{} remask threshold over $\{0.5,\ldots,0.9\}$. Point labels are the swept correction threshold. Horizontal dotted line: \TtoT{} best on that panel.}
    \label{fig:design_b_multitask}
\end{figure}

Fig.~\ref{fig:design_ab} in the main text reports the two orthogonal threshold sweeps on GSM8K under the paper-default \textsc{LowProb} threshold ($\tau_{\mathrm{lp}}{=}0.7$). Figs.~\ref{fig:design_a_multitask} and~\ref{fig:design_b_multitask} repeat the same protocols on four further full benchmarks (HumanEval+, MBPP+, C-EVAL, PIQA), keeping GSM8K for reference. Both figures plot accuracy against NFE/token with the swept threshold annotated on each point; \TtoM{} always uses $\tau_{\mathrm{lp}}{=}0.7$ (no $\tau_{\mathrm{lp}}{=}0.9$ leg).

Across Design~A (fill sweep), \TtoT{} saturates or slows while \TtoM{} raises the reachable ceiling on every task ($+1.06$ to $+3.05$). Across Design~B (edit/remask-threshold sweep at fixed fill $0.7$), making edits more aggressive never closes the gap: \TtoM{} remains above \TtoT{} on all five tasks. The GSM8K pattern is therefore not an outlier of one benchmark.

\section{Latency, Throughput, and Remasking Budget}
\label{app:latency}

Table~\ref{tab:main} reports NFE/token as exact algorithm-level forward counts; Table~\ref{tab:latency} adds wall-clock latency and throughput under the same paired protocols (A100-80GB, batch size 1). For medium and long outputs, latency generally increases with the forward count; short outputs show larger fixed-overhead deviations.

On C-EVAL, tightening the default caps from $(C_{\max},\rho_{\max})=(3,0.5)$ to $(1,0.25)$ gives $77.04$ at $0.86$ NFE/token, compared with the \TtoT{} baseline's $75.78$, thereby recovering an operating point below the AR reference $1.0$. This is the worked example for the tunable-budget statement in Section~\ref{sec:cost}; we do not apply the same budget tightening to the remaining tasks.

\begin{table}[!htbp]
\centering
\caption{Per-benchmark latency and throughput (LLaDA2.1-mini, paired task-standard protocols; A100-80GB, batch size 1). Len.\ and NFE/tok.\ match Table~\ref{tab:main} and are exact algorithm-level forward counts; latency and tok/s are measured from per-request serving logs.}
\label{tab:latency}
\vspace{2pt}
\scriptsize
\setlength{\tabcolsep}{3.5pt}
\begin{tabular}{lrrrrrr}
\toprule
\textbf{Benchmark} & \textbf{Len.\ T2T/T2M} & \textbf{NFE/tok.\ T2T} & \textbf{NFE/tok.\ T2M} & \textbf{$\times$} & \textbf{Latency T2T / tok/s} & \textbf{Latency T2M / tok/s} \\
\midrule
GSM8K & 296 / 292 & 0.26 & 0.58 & 2.2 & 0.92\,s / 321 & 1.99\,s / 147 \\
OlympiadBench & 4440 / 4427 & 0.26 & 0.62 & 2.4 & 12.9\,s / 345 & 27.6\,s / 159 \\
GSM-Plus & 335 / 344 & 0.32 & 0.84 & 2.6 & 1.16\,s / 289 & 3.07\,s / 112 \\
Omni-MATH & 5452 / 5426 & 0.27 & 0.66 & 2.4 & 13.7\,s / 397 & 36.3\,s / 150 \\
CMATH & 301 / 302 & 0.32 & 0.77 & 2.4 & 1.09\,s / 279 & 2.44\,s / 121 \\
\midrule
HumanEval+ & 353 / 335 & 0.14 & 0.29 & 2.0 & 0.50\,s / 711 & 1.09\,s / 307 \\
MBPP+ & 268 / 273 & 0.18 & 0.37 & 2.1 & 0.39\,s / 708 & 1.04\,s / 263 \\
\midrule
GPQA-Diamond & 4588 / 4629 & 0.41 & 1.12 & 2.7 & 18.0\,s / 256 & 50.9\,s / 92 \\
C-EVAL & 882 / 820 & 0.51 & 1.48 & 2.9 & 5.2\,s / 168 & 11.3\,s / 72 \\
MMLU-Pro & 1907 / 1921 & 0.37 & 0.97 & 2.6 & 6.5\,s / 275 & 18.2\,s / 105 \\
TriviaQA & 9 / 9 & 0.83 & 2.16 & 2.6 & 0.18\,s / 53 & 0.23\,s / 40 \\
\midrule
PIQA & 95 / 105 & 0.76 & 2.04 & 2.7 & 0.93\,s / 143 & 2.6\,s / 52 \\
OCNLI & 138 / 138 & 0.58 & 1.69 & 2.9 & 0.80\,s / 173 & 2.36\,s / 58 \\
DROP & 232 / 223 & 0.33 & 0.73 & 2.2 & 0.77\,s / 267 & 1.7\,s / 130 \\
BBH & 665 / 620 & 0.42 & 0.96 & 2.3 & 2.4\,s / 246 & 6.0\,s / 103 \\
HellaSwag & 251 / 250 & 0.55 & 1.50 & 2.7 & 1.4\,s / 174 & 4.0\,s / 63 \\
\midrule
IFEval & 461 / 482 & 0.70 & 2.17 & 3.1 & 2.9\,s / 161 & 9.6\,s / 51 \\
\bottomrule
\end{tabular}
\end{table}

\section{Trajectory-Level Mechanism Analysis}
\label{app:mechanism}

Per-step statistics on GSM8K show that most remasks leave the token unchanged: under the $\tau{=}0.9$ configuration, only $1.2\%$ of remasked positions refill with a different token. Under the default configuration, baseline-error-to-\TtoM{}-correct outcome flips (fixes) outnumber baseline-correct-to-\TtoM{}-error flips (breaks) $56$ to $35$; HumanEval+ contributes $8$ fixes against $2$ breaks and CMATH $25$ against $19$.

Repeated reopening matters despite this sparsity. Cutting the per-position budget to $C_{\max}{=}1$ reduces the GSM8K gain to $+0.23$ at $0.32$ NFE/token, suggesting that some positions are refilled before their neighborhoods have stabilized and must be reopened later.

An immediate-refill control separates masking from cross-step deferral. With the same \textsc{LowProb} detector, immediate masked-context refill reaches $89.99$, between immediate replacement ($89.39$) and deferred refill ($91.13$). The adjacent differences are not individually significant (exact McNemar $p{=}.51$ and $p{=}.13$), but the ordering is consistent with both masked-context re-prediction and subsequent context evolution contributing to the full effect. Only $8.5\%$ of immediate refills change the original token.

\section{CORE Baseline Port}
\label{app:core}

The CORE rows of Table~\ref{tab:mechanisms} run CORE's published detection and action~\citep{zhai2026core} on LLaDA2.1-mini rather than on its native LLaDA-8B-Base. The choice is deliberate: LLaDA-8B-Base has no correction training, and our own rule has little effect there ($+0.08$; Table~\ref{tab:crossmodel}), so a comparison on that model would put neither method in its domain.

The port keeps CORE's default hyperparameters as released: replacement gate $0.3$, one revision per pass, and re-mask interval $8$ (also swept to $2$). The only setting that cannot carry over verbatim is the candidate count: CORE's default of $32$ candidates is defined against a $512$-token block and would mask our $32$-token block entirely. The block-proportional equivalent is $2$; we allocate $4$ candidates.

One mechanistic difference matters for interpreting the comparison. CORE's mask is \emph{transient}: the flagged token is resampled within the same step from the auxiliary perturbed pass, which makes the action closer to a context-guided replace. \TtoM{} instead \emph{defers} the commitment across steps, holding the position open until its neighbourhood has converged further. The cross-model contrast in Section~\ref{sec:crossmodel} therefore concerns deferred refilling, whereas CORE resolves its transient mask within the same step. On LLaDA2.1-mini, CORE improves pure \MtoT{} decoding by $+0.53$, compared with the $+1.29$ reported on its native LLaDA-8B-Base setting; porting effects cannot be excluded.

\section{Detector Task-Specificity}
\label{app:logitdiff}

The detector family is not interchangeable across task types. Extending \textsc{LogitDiff} ($\tau_{\mathrm{ld}}{=}0.1$) beyond GSM8K to six further benchmarks at the same caps as \textsc{LowProb} ($C_{\max}{=}3$, $\rho_{\max}{=}0.5$) gives a one-sided split (Table~\ref{tab:logitdiff}): \textsc{LogitDiff} wins on the long-form math suites, ties on GSM-Plus, and loses on all four non-math tasks. We therefore retain \textsc{LowProb} as the task-general default.

\begin{table}[!htbp]
\centering
\caption{\textsc{LowProb} vs.\ \textsc{LogitDiff} ($\tau_{\mathrm{ld}}{=}0.1$) at matched caps. GSM8K uses the ablation budget; other rows follow Table~\ref{tab:main}.}
\label{tab:logitdiff}
\vspace{2pt}
\small
\begin{tabular}{lccc}
\toprule
\textbf{Task} & \textbf{\textsc{LowProb}} & \textbf{\textsc{LogitDiff}} & \textbf{$\Delta$} \\
\midrule
CMATH      & 92.35 & \textbf{93.44} & $+1.09$ \\
GSM8K      & 91.13 & \textbf{91.51} & $+0.38$ \\
GSM-Plus   & \textbf{81.79} & 81.77 & $-0.02$ \\
GPQA       & \textbf{45.96} & 42.42 & $-3.54$ \\
IFEval     & \textbf{84.47} & 82.99 & $-1.48$ \\
HumanEval+ & \textbf{80.49} & 79.27 & $-1.22$ \\
MBPP+      & \textbf{70.11} & 68.78 & $-1.33$ \\
\bottomrule
\end{tabular}
\end{table}

\section{Baseline Reproduction Quality}
\label{app:repro}

All \TtoT{} baseline numbers in this paper are our own reproduction with the open-source SGLang dLLM stack under the protocols of Section~\ref{sec:setup}. Table~\ref{tab:repro} compares them against the reported Q~Mode values on the sixteen comparable benchmarks; GSM8K is not reported for LLaDA2.1-mini in the technical report. All \TtoT{}--\TtoM{} comparisons use our reproduction under matched engines, prompts, budgets, and scorers.

\begin{table}[!htbp]
\centering
\caption{Reproduced \TtoT{} vs.\ LLaDA2.1-mini Q~Mode in the technical report~\citep{nie2026llada21}. Positive $\Delta$ = our reproduction is higher.}
\label{tab:repro}
\vspace{4pt}
\small
\begin{tabular}{lccc}
\toprule
\textbf{Benchmark} & \textbf{Reported Q} & \textbf{Ours (\TtoT{})} & \textbf{$\Delta$} \\
\midrule
HellaSwag      & 76.19 & 79.50 & $+3.31$ \\
DROP           & 82.37 & 84.14 & $+1.77$ \\
BBH            & 80.58 & 81.92 & $+1.34$ \\
IFEval (prompt-strict) & 83.18 & 83.36 & $+0.18$ \\
OCNLI          & 61.59 & 60.78 & $-0.81$ \\
TriviaQA       & 54.24 & 53.18 & $-1.06$ \\
MMLU-Pro       & 64.84 & 63.41 & $-1.43$ \\
\midrule
CMATH          & 94.99 & 92.26 & $-2.73$ \\
C-EVAL         & 78.59 & 75.63 & $-2.96$ \\
Omni-MATH      & 43.60 & 40.31 & $-3.29$ \\
GSM-Plus (full) & 86.55 & 81.37 & $-5.18$ \\
MBPP+          & 74.07 & 68.78 & $-5.29$ \\
PIQA           & 86.89 & 80.90 & $-5.99$ \\
HumanEval+     & 82.93 & 76.83 & $-6.10$ \\
OlympiadBench  & 66.67 & 58.16 & $-8.51$ \\
GPQA-Diamond   & 53.28 & 43.94 & $-9.34$ \\
\bottomrule
\end{tabular}
\end{table}

\paragraph{AIME 2025.} We exclude this 30-problem benchmark~\citep{aime2025} from the main suite because it is too small to support a stable paired comparison; all claims rely on the 17 full-size benchmarks of Table~\ref{tab:main}.

\section{Benchmark Details}
\label{app:benchmarks}

Table~\ref{tab:benchmark_details} lists all benchmarks used in our evaluation with dataset details.
The math suite comprises GSM8K~\citep{cobbe2021gsm8k}, GSM-Plus~\citep{li2024gsm_plus}, CMATH~\citep{wei2023cmath}, OlympiadBench~\citep{he2024olympiadbench}, and Omni-MATH~\citep{gao2024omnimath}.
Code evaluation uses HumanEval and MBPP under EvalPlus~\citep{chen2021humaneval,liu2023evalplus}.
Knowledge evaluation covers GPQA~\citep{rein2024gpqa}, C-EVAL~\citep{huang2023ceval}, MMLU-Pro~\citep{wang2024mmlu_pro}, and TriviaQA~\citep{joshi2017triviaqa}.
The remaining suite includes BBH~\citep{suzgun2023bbh}, DROP~\citep{dua2019drop}, HellaSwag~\citep{zellers2019hellaswag}, PIQA~\citep{bisk2020piqa}, OCNLI~\citep{hu2020ocnli}, and IFEval~\citep{zhou2023ifeval}.
IFEval uses the prompt-strict Google checker; because its vendored instruction constructors draw random defaults when optional arguments are absent, we reset Python's RNG to seed 0 before independently rescoring each decoding leg, making the comparison deterministic and matched.

\begin{table}[!htbp]
\centering
\caption{Benchmark details. Generation budget $16{,}384$ except TriviaQA (50).}
\label{tab:benchmark_details}
\footnotesize
\begin{tabular}{llcclc}
\toprule
\textbf{Category} & \textbf{Benchmark} & \textbf{Split} & \textbf{Samples} & \textbf{Prompt} & \textbf{Metric / Scorer} \\
\midrule
\multirow{5}{*}{Math}
& GSM8K          & test  & 1,319   & 4-shot (dInfer)        & EM (numeric) \\
& OlympiadBench  & maths & 674     & 0-shot, boxed          & EM (numeric) \\
& GSM-Plus       & test  & 10,552  & 4-shot + None protocol & EM (numeric) \\
& Omni-MATH      & test  & 4,428   & 0-shot, boxed          & EM (numeric) \\
& CMATH          & test  & 1,098   & 0-shot, boxed          & EM (numeric) \\
\midrule
\multirow{2}{*}{Code}
& HumanEval+     & test  & 164     & evalplus instruct      & pass@1 (evalplus) \\
& MBPP+          & test  & 378     & evalplus instruct      & pass@1 (evalplus) \\
\midrule
\multirow{4}{*}{Science/Knowl.}
& GPQA-Diamond   & test  & 198     & 0-shot, option rotation & Accuracy    \\
& C-EVAL         & val   & 1,346   & 0-shot MCQ (zh)        & Accuracy    \\
& MMLU-Pro       & test  & 12,032  & 0-shot CoT             & Accuracy    \\
& TriviaQA       & val   & 17,944  & 5-shot, short answer   & EM (alias match) \\
\midrule
\multirow{5}{*}{Reasoning}
& PIQA           & val   & 1,838   & 0-shot MCQ             & Accuracy    \\
& OCNLI          & val   & 2,950   & 0-shot (zh NLI)        & Accuracy    \\
& DROP           & val   & 9,536   & 3-shot (simple-evals)  & EM (fuzzy match) \\
& BBH            & test  & 6,511   & 3-shot CoT (official)  & Accuracy    \\
& HellaSwag      & val   & 10,042  & 0-shot MCQ             & Accuracy    \\
\midrule
Instruction
& IFEval         & official & 541  & raw prompt             & Prompt-strict (Google checker) \\
\bottomrule
\end{tabular}
\end{table}

\end{document}